\documentclass[letterpaper]{article} %
\usepackage{aaai24}  %
\usepackage{times}  %
\usepackage{helvet}  %
\usepackage{courier}  %
\usepackage[hyphens]{url}  %
\usepackage{graphicx} %
\usepackage{natbib}  %
\usepackage{caption} %
\usepackage{algorithm}
\usepackage{algorithmic}

\usepackage{amsmath} 
\usepackage{amssymb} 
\usepackage{newfloat}
\usepackage{listings}
\DeclareCaptionStyle{ruled}{labelfont=normalfont,labelsep=colon,strut=off} %
\floatstyle{ruled}
\newfloat{listing}{tb}{lst}{}
\floatname{listing}{Listing}
\title{Deep Linear Array Pushbroom Image Restoration: A Degradation Pipeline and Jitter-Aware Restoration Network}
\author{
    Zida Chen\textsuperscript{\rm 1}\equalcontrib,
    Ziran Zhang\textsuperscript{\rm 1,\rm 2}\equalcontrib,
    Haoying Li\textsuperscript{\rm 1},
    Menghao Li\textsuperscript{\rm 1}, \\
    Yueting Chen\textsuperscript{\rm 1},
    Qi Li\textsuperscript{\rm 1},
    Huajun Feng\textsuperscript{\rm 1},
    Zhihai Xu\textsuperscript{\rm 1},
    Shiqi Chen\textsuperscript{\rm 1}\thanks{Corresponding author}
}
\affiliations{
    \textsuperscript{\rm 1}State Key Laboratory of Extreme Photonics and Instrumentation, Zhejiang University\\
    \textsuperscript{\rm 2}Shanghai Artificial Intelligence Laboratory\\
    \{zd\_chen, naturezhanghn, lhaoying, limh, chenyt, liqi, fenghj, xuzh, chenshiqi\}@zju.edu.cn \\

}

\begin{document}

\maketitle

\begin{abstract}

Linear Array Pushbroom (LAP) imaging technology is widely used in the realm of remote sensing. However, images acquired through LAP always suffer from distortion and blur because of camera jitter. Traditional methods for restoring LAP images, such as algorithms estimating the point spread function (PSF), exhibit limited performance. To tackle this issue, we propose a Jitter-Aware Restoration Network (JARNet), to remove the distortion and blur in two stages. In the first stage, we formulate an Optical Flow Correction (OFC) block to refine the optical flow of the degraded LAP images, resulting in pre-corrected images where most of the distortions are alleviated. In the second stage, for further enhancement of the pre-corrected images, we integrate two jitter-aware techniques within the Spatial and Frequency Residual (SFRes) block: 1) introducing Coordinate Attention (CoA) to the SFRes block in order to capture the jitter state in orthogonal direction; 2) manipulating image features in both spatial and frequency domains to leverage local and global priors. Additionally, we develop a data synthesis pipeline, which applies Continue Dynamic Shooting Model (CDSM) to simulate realistic degradation in LAP images. Both the proposed JARNet and LAP image synthesis pipeline establish a foundation for addressing this intricate challenge. Extensive experiments demonstrate that the proposed two-stage method outperforms state-of-the-art image restoration models. Code is available at https://github.com/JHW2000/JARNet.

\end{abstract}

\section{Introduction}

\begin{figure}
\centering
\includegraphics[width=0.47\textwidth]{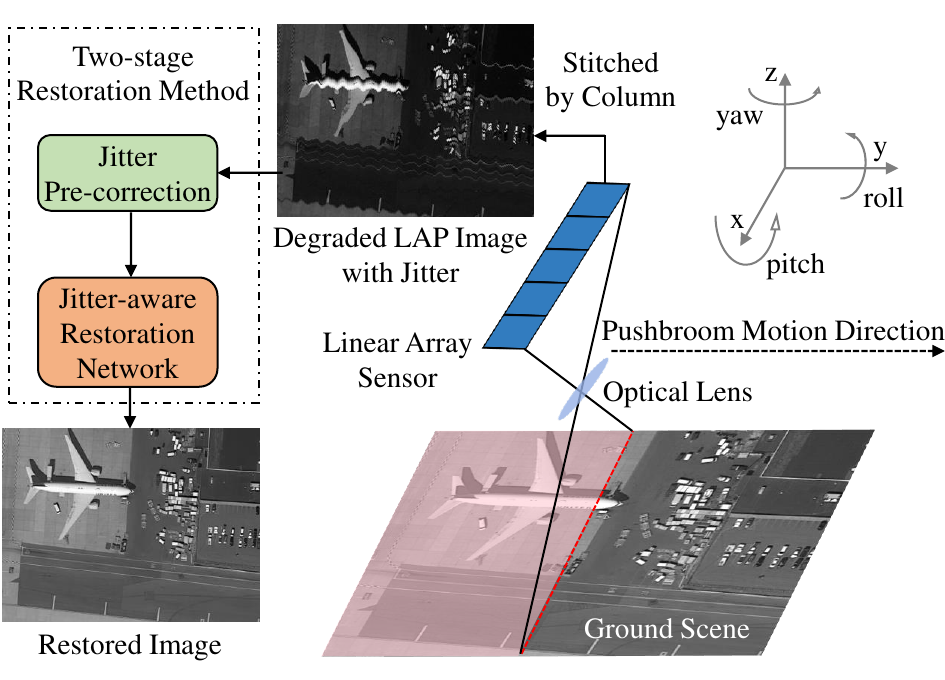}
\caption{Illustration of our main idea: The stitched LAP image suffers from distortion and blur caused by camera jitter. To restore degraded LAP image, we propose a jitter pre-correction and enhancement method in a two-stage manner. We also design an image synthesis pipeline for training data acquisition. Finally, our proposed JARNet outperforms state-of-the-art methods on our LAP dataset.}
\label{fig:camera}
\end{figure}

Linear array cameras are widely employed in remote sensing for high-resolution optical imaging on the earth \cite{cui2023optical, wang2018development}. As shown in Fig. \ref{fig:camera}, a linear array camera contains an array of sensors arranged in a straight line. The linear array sensor captures images at ground scene while the whole camera moves along the pushbroom motion direction. This movement is similar to how a broom sweeps forward. Pixels imaged at different moments are stitched together to generate a LAP image. Due to adjustments in camera attitude and periodic movement \cite{iwasaki2011detection}, inevitable jitter arises in LAP imaging. This leads to pixel displacements of varying magnitudes, thereby inducing distortion and blur within LAP images. Low-frequency jitter causes image distortion effect, while high-frequency jitter leads to blur effect \cite{pan2020jitter}. In practice, roll jitter causes pixel offset in the cross-track direction, and pitch jitter causes pixel offset in the along-track direction. Jitter offsets in the along-track direction are much smaller than that in the cross-track direction, while jitter in the yaw direction is tiny enough to be neglected \cite{wang2017image}.

Many efforts have been made to remove distortion and blur in LAP images. Their efforts are directed towards acquiring more precise jitter curves, either directly measuring from high-resolution equipment \cite{pan2020jitter} or indirectly predicting from degraded images \cite{wang2021jitter}. However, the former is often hindered by the scarcity of high-precision equipment, while the latter often falls short in terms of predicting accuracy. Over the years, deep learning has demonstrated outstanding performance in many image restoration tasks due to its powerful data modeling and generation capabilities. However, mainstream image restoration methods \cite{cho2021rethinking, chen2021hinet, zamir2022restormer, wang2022uformer} are not well-adopted in LAP image reconstruction due to the lack of jitter priors. As shown in Fig. \ref{fig:main_result}, we test seven mainstream methods and they struggle to restore sufficient details of farmland and city buildings, showing their limitation in LAP image restoration task. 

Another obstacle is the availability of data that comes with distortion and blur by jitter. As far as we are concerned, there is no customized LAP image dataset. Many studies \cite{zhaoxiang2019attitude, wang2021jitter} use several sinusoidal components to simulate jitter effect in LAP image \cite{wulich1987image}. However, jitter in the real world is more complicated. So how to establish an efficient data synthesis pipeline remains a problem.

\begin{figure*}[ht]
\centering
\includegraphics[width=\textwidth]{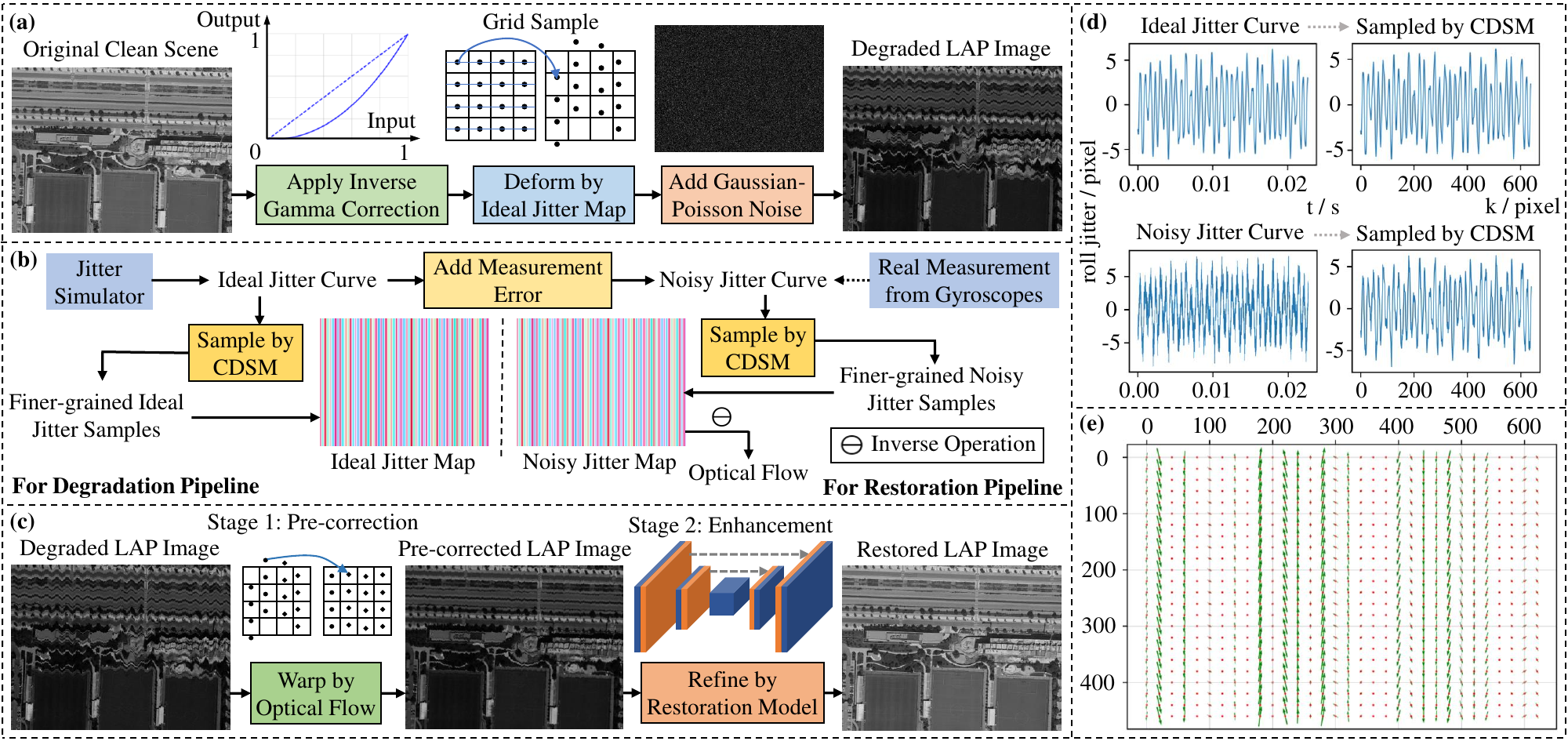}
\caption{Overall degradation and restoration pipeline for LAP images. (a) Proposed LAP image degradation pipeline. (b) Proposed CDSM-based jitter map generating procedure. (c) Proposed two-stage restoration pipeline for LAP images. (d) Visualization of jitter curve in (b). From upper left to lower right: ideal jitter curve, the average of finer-grained ideal jitter samples from CDSM, noisy jitter curve, and the average of finer-grained noisy jitter samples from CDSM. For simplicity, we only display jitter curve in the roll direction. (e) Visualization of optical flow from noisy jitter map.}
\label{fig:pipeline}
\end{figure*}

In this paper, we tackle the challenge of limited data availability for LAP images and propose a \textbf{novel restoration pipeline} for LAP image recovery. Inspired by Continue Dynamic Shooting Model (CDSM) which provides a finer-grained sampling strategy \cite{pan2020jitter}, we propose a \textbf{novel LAP image degradation pipeline} on CDSM-based jitter model, which simulates distortion and blur. We generate sufficient degraded LAP images from public dataset DOTA-v1.0 \cite{Xia_2018_CVPR} for building our LAP dataset.

In order to utilize jitter prior and achieve better restoration performance, we propose JARNet, a \textbf{jitter-aware restoration network} based on the two-stage restoration strategy. In the first stage, we design an Optical Flow Correction (OFC) block to refine optical flow from jitter prior. Then we warp the degraded LAP image by the refined optical flow to make precise pre-correction, which removes most of the distortion. In the second stage, we design a Spatial and Frequency Residual (SFRes) block, which integrates two jitter-aware techniques of coordinate attention \cite{hou2021coordinate} (CoA) block and frequency branch \cite{mao2023intriguing}. The CoA block in spatial branch captures jitter state in orthogonal direction. The frequency branch parallel to the spatial branch extracts both low and high-frequency jitter, guiding the overall removal of distortion and blur. Compared with state-of-the-art methods, our proposed JARNet achieves competitive performance in LAP image restoration task. In summary, our contributions are outlined as follows:

\begin{itemize}
\item We develop a novel LAP image synthesis pipeline with CDSM integration, which achieves a finer-grained level of simulation fidelity and boosts restoration performance.
\item We propose the first jitter-aware restoration network, employing optical flow correction and two jitter-aware techniques to utilize both spatial and frequency information.
\item Extensive experiments show that our method achieves superior results (+ 1.28dB in PSNR) compared with state-of-the-art methods on our LAP dataset.
\end{itemize}

\section{Related Work}
\subsection{LAP Image Restoration for Remote Sensing}

Many efforts have been made to remove distortion and blur effects in LAP images. For distortion caused by low-frequency jitter, many researchers improve the accuracy of jitter detection for image resampling. While high-precision attitude sensors \cite{tang2014high} directly measure the attitude of linear array cameras, the feasibility is constrained by its economic viability. Some works focus on leveraging parallax imaging systems across distinct spectral bands within multispectral images for indirect prediction of jitter state \cite{hu2018high}. Nevertheless, such strategies are often rendered inapplicable within the context of panchromatic LAP images. Furthermore, solutions based on deep learning have also been explored to predict jitter curves \cite{zhaoxiang2019attitude}. However, these approaches only work within a limited range of jitter amplitude. 

For blur effects caused by high-frequency jitter, traditional techniques are divided into blind and non-blind restoration methods. The former targets scenarios where blur kernel remains unknown, while the latter undertakes algorithms based on the point spread function (PSF) \cite{vimal2019mixture,chen2013estimating}. Most studies usually apply the same PSF to the entire image, because jitter is regarded as uniform within a short imaging interval \cite{pan2020jitter}. However, this assumption is invalid for high-frequency jitter. 

As the real LAP image data is rare, researchers often \cite{zhaoxiang2019attitude, wang2021jitter} simulate jitter effects by using multiple sinusoidal components \cite{wulich1987image}. However, jitter in the real world is more intricate. Consequently, establishing an effective data synthesis pipeline remains a challenge. In this paper, we establish a CDSM-based LAP image degradation pipeline to acquire sufficient LAP data. We utilize deep learning technology and propose JARNet, to process distortion and blur in a two-stage restoration manner. We remove most of the distortions with jitter prior processed by CDSM in the first stage. Then we deal with the rest of distortion and blur effect by restoration network in the second stage.

\subsection{Learning-based Single Image Restoration}

Deep learning has emerged as a powerful tool for learning data-driven models end-to-end, especially in low-level vision tasks (e.g. image restoration). Many methods have achieved remarkable performance on public degraded datasets \cite{nah2017deep, abdelhamed2018high, li2023real}, proving their potential in natural image restoration. These methods often employ multi-scale architectures. For example, MIMO-UNet \cite{cho2021rethinking} utilizes multi-scale inputs and outputs, HINet \cite{chen2021hinet} applies a two-stage UNet, and NAFNet \cite{chen2022simple} leverages a simplified UNet backbone. Some transformer-based approaches utilize self-attention mechanism and reduce time complexity of vanilla vision transformer \cite{dosovitskiy2021an}, such as Uformer \cite{wang2022uformer}, Restormer \cite{zamir2022restormer} and Stripformer \cite{tsai2022stripformer}. However, such mainstream image restoration methods do not work well in LAP image restoration task due to the domain gap between LAP and natural imaging scenes. With the refined optical flow from OFC block and two jitter-aware techniques of CoA block and frequency branch, our JARNet can capture jitter state in orthogonal direction and extract jitter at different frequencies, which is well-adopt in LAP image restoration.

\section{Proposed Method}

We first introduce the principle of LAP jitter with the CDSM-based dense sampling strategy. On this basis, we propose an image synthesis pipeline for LAP image degradation. Finally, we present the jitter-aware image restoration network, JARNet, in a two-stage restoration manner.

\subsection{CDSM-Based Jitter Model}

For linear array sensors in the remote sensing field, the main factor of degradation is the image deformation and blur effect caused by the camera jitter. The jitter can be described by a series of time-varying sinusoidal functions \cite{wulich1987image} in Eq. \ref{eq: jitter}:
\begin{equation}
\phi_{d}(t)=\sum_{i}^{N}A_{i}\sin \left(2\pi f_{i}t+\varphi_{i}\right),
\label{eq: jitter}
\end{equation}
where $\phi _{d}$ is a time-varying jitter angle curve in a certain direction, $A$, $f$ and $\varphi$ are jitter amplitude, jitter frequency, and initial random phase, respectively. $N$ means the number of sinusoidal components. Due to tiny $\phi _{d}$, the pixel offset on the sensor caused by jitter can be approximated by $J_{d}=\phi _{d}\frac{f}{\mu}$, where $J_{d}$ is the number of pixel offset in a certain direction, $f$ is focal length of optical system, and $u$ is pixel size of linear array sensor. Proportional to $\phi _{d}$, $J_{d}$ consists of various sinusoidal components. As shown in Fig. \ref{fig:pipeline} (b), (d), we obtain ideal jitter curve $J_{d}$ for LAP image degradation pipeline by jitter simulator in Eq. \ref{eq: jitter}. The real jitter curve measured by gyroscopes is noisy that contains much measurement error. We denote it as noisy jitter curve. However, it is difficult to obtain sufficient noisy jitter curves in the real world. Therefore, we simulate noisy jitter curve for LAP two-stage restoration pipeline by adding measurement error to the ideal jitter curve.

It is worth noting that ideal jitter curve from jitter simulator simulates image distortion with only tiny blur effect. And correcting deformed LAP image directly by noisy jitter curve is imprecise. So we introduce CDSM into our jitter model to address these issues, which will be discussed in Section \ref{section:A} and Section \ref{section:B}. CDSM provides a denser sampling strategy \cite{pan2020jitter}. Without CDSM, we sample jitter curve at the time interval of imaging $\tau$. In other words, in Eq. \ref{eq: jitter}, $t \subseteq k\tau$, where $k$ is column pixel index in LAP image. With the application of CDSM, we utilize a shorter time interval for finer-grained sampling. Derived from Eq. \ref{eq: jitter}, The CDSM-based jitter model is depicted by Eq. \ref{eq: CDSM_sub}:
\begin{equation}
J_{d}^{sub}(t, m)=\sum_{i}^{N} A_{i}^{'} \sin \left(2 \pi f_{i}(t+\frac{m}{M} \tau)+\varphi_{i}\right),
\label{eq: CDSM_sub}
\end{equation}
where $J_{d}^{sub}$ is the subdivision jitter curve, $A_{i}^{'}=A_{i}\frac{f}{\mu}$, $M$ is subdivision number, $m$ is subdivision index, $\tau$ is time interval of imaging. With CDSM, the sample time interval becomes $\frac{\tau}{M}$, achieving finer-grained sampling. In the upper right and lower right of Fig. \ref{fig:pipeline} (d), we visualize the effect of CDSM by averaging finer-grained jitter curves in Eq. \ref{eq: CDSM}:
\begin{equation}
J_{d}^{CDSM} (k)=\frac{1}{M} \sum_{m}^{M} J_{d}^{sub}(k\tau, m),
\label{eq: CDSM}
\end{equation}
where $J_{d}^{CDSM}$ is jitter curve processed by CDSM. Compared to their original state, the ideal jitter curve shows only minor changes, while the noisy jitter curve is smoothed.

\subsection{LAP Image Degradation Pipeline}
\label{section:A}

\begin{figure}
\centering
\includegraphics[width=0.47\textwidth]{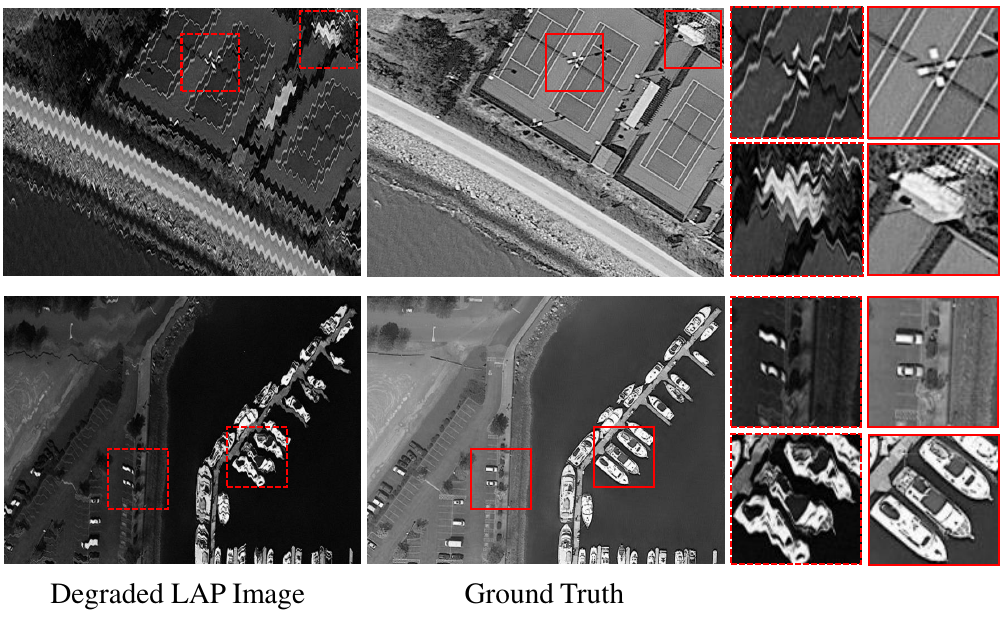}
\caption{Examples of our LAP image dataset. The $1^{st}$ and $3^{rd}$ columns are simulated LAP images from our proposed degradation pipeline. The $2^{nd}$ and $4^{th}$ columns are the corresponding original clean scene.
}
\label{fig:dataset}
\end{figure}

Fig. \ref{fig:pipeline} (a) illustrates the pipeline of generating degraded LAP images from original clean scene. Firstly, we apply inverse gamma correction for original clean image to obtain the energy representation of the scene, as the sensor noise follows Gaussian-Poisson distribution in this domain. Next, we deform the image by ideal jitter map. Finally, Gaussian-Poisson noise is added to the deformed image. 

Ideal jitter map determines the amount of jitter offset for each pixel in LAP image. Given a LAP image with shape of $(H, W, 1)$, where the width direction is the pushbroom direction, we establish a sampling time array $t \subseteq \left \{ \tau,2\tau,...,W\tau \right \} $. According to Eq. \ref{eq: CDSM_sub}, we calculate a finer-grained ideal jitter sample in a certain direction from CDSM with a shape of $(W, 1)$, denoted as $\widetilde J_{d}^{sub}$. The height direction represents the direction of the linear array sensor. Since all the pixels are imaged at the same time, their jitter states remain consistent. So we simply duplicate the jitter sample along the height direction to a shape of $(H, W, 1)$. Subsequently, we concatenate jitter from roll and pitch direction, resulting in an ideal jitter map with a shape of $(H, W, 2)$. Eq. \ref{jitter map} shows the process above.
\begin{equation}
J^{sub}(m)=\boldsymbol{C}\bigl(\boldsymbol{D}(\widetilde J_{roll}^{sub}(m), h), \boldsymbol{D}(\widetilde J_{pitch}^{sub}(m), h)\bigr),
\label{jitter map}
\end{equation}
where $J^{sub}$ is the $m^{th}$ subdivided ideal jitter map. $\boldsymbol{D}$ is duplicate operation. $h$ represents the height direction. $\boldsymbol{C}$ is concatenate operation. Finally, we make a grid sample to deform LAP image by several subdivision ideal jitter maps. Since the pixel offset may not be an integer, we apply bilinear interpolation during resampling. The LAP image degradation pipeline is depicted by Eq. \ref{degraded}:
\begin{equation}
I^{lq} =\frac{1}{M}\sum_{m}^{M}\boldsymbol G\bigl((I^{gt})^\gamma, J^{sub}(m)\bigr)+n,
\label{degraded}
\end{equation}
where $I^{lq}$ is the degraded LAP image, $I^{gt}$ is the original image, $\gamma$ is inverse gamma correction coefficient, $\boldsymbol{G}$ denotes grid sample operation, and $n$ is Gaussian-Poisson noise. Through averaging deformed LAP images from ideal jitter maps with different subdivision indexes, we successfully introduce blur effect into degraded LAP image thanks to CDSM-based jitter model.

Finally, we establish our LAP dataset through our proposed degradation pipeline, which contains degraded-clean LAP image pairs. We demonstrate two LAP image pairs and their zoom-in details in Fig. \ref{fig:dataset}.

\subsection{Jitter-Aware Restoration Network}
\label{section:B}
To restore the LAP images from the pixel displacement and blur caused by camera jitter, we propose a jitter-aware restoration network, JARNet, which restores LAP images in two stages, as shown in Fig. \ref{fig:pipeline} (c). The first stage is called the pre-correction stage, where most of the distortion in degraded LAP image is removed by optical flow. Noticeably, we introduce Optical Flow Correction (OFC) block in JARNet to enable more precise warping and effective distortion removal. The second stage is called the enhancement stage, where we employ a Spatial and Frequency Residual (SFRes) block in the U-shaped network to further enhance the pre-corrected LAP image. The two-stage restoration strategy offers us an effective approach to enhance the performance of state-of-the-art methods in LAP image restoration task. We will introduce the two stages, the proposed jitter-aware techniques, and losses in the following paragraphs.

\begin{figure*}[ht]
\centering
\includegraphics[width=\textwidth]{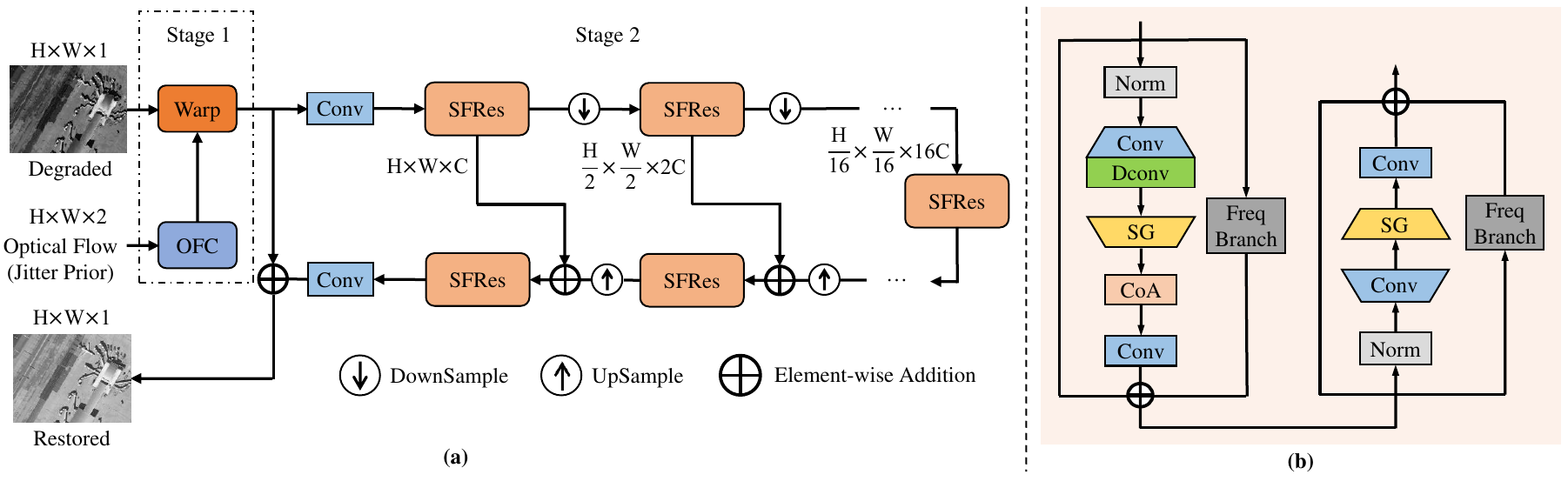}
\caption{Architecture of JARNet for LAP image restoration. (a) Overview of our two-stage JARNet. (b) Details of SFRes block. `Norm' means layer normalization. `Dconv' means depthwise convolution. `SG' means SimpleGate \cite{chen2022simple}. `CoA' means coordinate attention \cite{hou2021coordinate}. `Freq Branch' means frequency branch \cite{mao2023intriguing}.}
\label{fig:network_arch}
\end{figure*}

\begin{figure}
\centering
\includegraphics[width=0.47\textwidth]{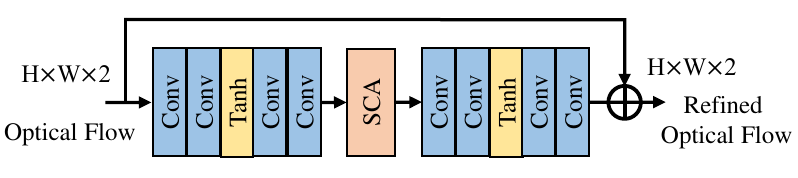}
\caption{Details of our proposed Optical Flow Correction (OFC) block. `Tanh' means hyperbolic tangent activation function. `SCA' means Simplified Channel Attention \cite{chen2022simple}.}
\label{fig:OFC}
\end{figure}

\noindent \textbf{Pre-correction Stage and OFC Block} \indent
Since jitter state provides degradation information, we attempt to use it as prior to warp the distorted LAP images in the pre-correction stage. Specifically, we simulate several subdivided noisy jitter maps from noisy jitter curve as shown in Fig. \ref{fig:pipeline} (b), whose process is similar to ideal jitter map. Then we average all the subdivided noisy jitter maps, assisting in smoothing the noise thanks to CDSM-based jitter model. Subsequently, the degraded LAP image is warped by optical flow in Eq. \ref{warp}, which is approximated as the inverse of noisy jitter map:
\begin{equation}
\begin{split}
I^{warp} &=\boldsymbol G(I^{lq}, \omega) \approx \boldsymbol G(I^{lq}, -J_{noisy})\\
&=\boldsymbol G\bigl(I^{lq}, -\frac{1}{M}\sum_{m}^{M}J^{sub}_{noisy}(m)\bigr),
\label{warp}
\end{split}
\end{equation}
where $I^{warp}$ is the pre-corrected image, most of whose distortions are removed. $\omega$ denotes optical flow, and $J_{noisy}$ denotes noisy jitter map. Fig. \ref{fig:pipeline} (e) shows the optical flow map with shape of $(H, W, 2)$. The vector at each pixel represents direction and relative magnitude of pixel offset. 

Instead of warping the degraded image by original optical flow, we refine the optical flow by our proposed Optical Flow Correction (OFC) block. As shown in Fig. \ref{fig:OFC}, the OFC block has a shallow attention-based convolution architecture and employs the SCA module \cite{chen2022simple} to reweight feature map of optical flow between roll and pitch directions, which assists better embedding of jitter prior. Then we utilize the refined optical flow to warp the degraded image.

\noindent \textbf{Enhancement Stage and SFRes Block} \indent
Since optical flow is calculated approximately, there still remains little distortion and blur in pre-corrected LAP image. In the enhancement stage, we further improve the LAP image quality by a U-shaped network with Spatial and Frequency Residual (SFRes) blocks as key blocks, as illustrated in Eq. \ref{refine}: 
\begin{equation}
I^{R} =\boldsymbol{N}(I^{warp}; \Theta),
\label{refine}
\end{equation}
where $I^{R}$ is restored image. $\boldsymbol{N}$ is restoration network. $\Theta$ is parameter set in image restoration network.
As shown in Fig. \ref{fig:network_arch}, the enhancement stage adopts a U-shaped network with skip connections between the encoder and the decoder. After the initial convolution layer, each level of encoder and decoder uses a series of SFRes blocks, inspired by the design of NAFNet block \cite{chen2022simple}. The SFRes block includes a spatial branch, a frequency branch, and a residual path, taking advantage of both local and global information. In both of the two kinds of branches, we adopt jitter-aware techniques to further improve the LAP image quality.

The spatial branch of SFRes block employs coordinate attention (CoA) \cite{hou2021coordinate} block to reweight feature map in the vertical and horizontal directions of LAP image. With the input feature shape of $(H, W, C)$, CoA outputs two attention maps with shapes of $(H, 1, C)$ and $(1, W, C)$, respectively. Then element-wise multiplication is applied between the input feature map and attention maps. CoA block helps capture the jitter state in orthogonal directions for better LAP image restoration.

The frequency branch \cite{mao2023intriguing} in the SFRes applies fast Fourier transform algorithm to convert the feature map to the frequency domain. Each pixel of feature map in frequency branch contains global information of LAP image, which not only extracts both low and high-frequency jitter but also assists in restoring high-frequency details. 

\noindent \textbf{Losses} \indent
To supervise the training of JARNet, we apply a restoration loss $\mathcal{L}_{res}$ and an optical flow loss $\mathcal{L}_{flow}$ in the total loss function $\mathcal{L}_{total}$, as illustrated in Eq. \ref{total loss}:
\begin{equation}
\mathcal{L}_{total}  = \mathcal{L}_{res}  + \lambda_1 \cdot \mathcal{L}_{flow},
\label{total loss}
\end{equation}
where $\lambda_1=0.1$.
The refined optical flow $\omega^{R}$ is supervised by noise-free optical flow $\omega^{gt}$, where $\omega^{gt} \approx -\frac{1}{M}\sum_{m}^{M}J^{sub}(m)$. We define the flow loss in Eq. \ref{eq:flow loss}:
\begin{equation}
\mathcal{L}_{flow} = \mathcal{L}_{1}(\omega^{R}, \omega^{gt}),
\label{eq:flow loss}
\end{equation}
where $\mathcal{L}_{1}$ is mean absolute error loss function. $\mathcal{L}_{flow}$ is utilized to constrain the optimization of the optical flow. We train OFC block as part of JARNet in a joint end-to-end manner. We apply a restoration loss to supervise the training of JARNet between
$I^{R}$ and $I^{gt}$ in Eq. \ref{restore loss}:
\begin{equation}
\mathcal{L}_{res}  = \mathcal{L}_{1}  + \lambda_2 \cdot \mathcal{L}_{percep} + \lambda_3 \cdot \mathcal{L}_{fft},
\label{restore loss}
\end{equation}
where $\lambda_2=10^{-4}$ and $\lambda_3=0.1$. $\mathcal{L}_{res}$ contains 3 components, mean absolute error loss, perceptual loss \cite{johnson2016perceptual}, and FFT loss \cite{cho2021rethinking}.

\section{Experiments}

\begin{figure*}[ht]
\centering
\includegraphics[width=\textwidth]{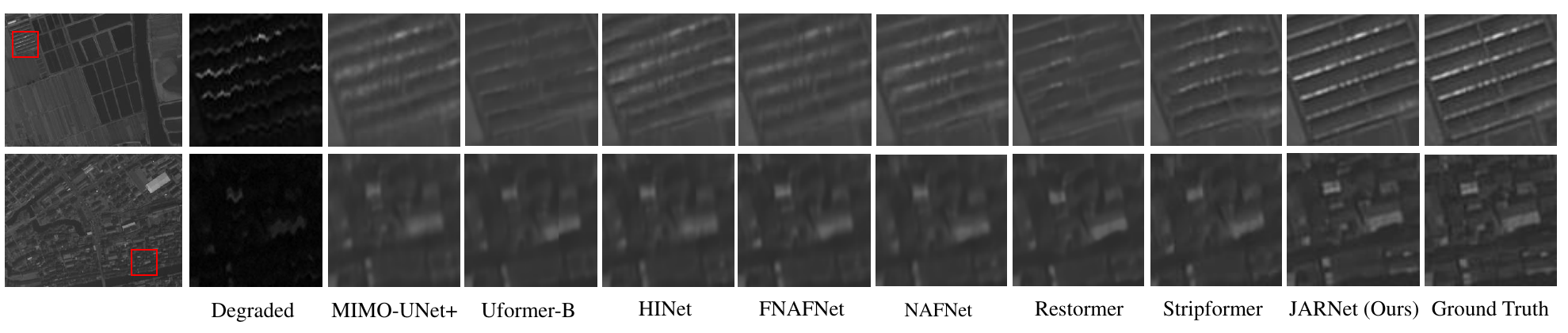}
\caption{Visual comparison of different restoration methods on our LAP dataset.}
\label{fig:main_result}
\end{figure*}

\subsection{Dataset}

We utilize DOTA-v1.0 dataset \cite{Xia_2018_CVPR} as our source of clean data, which is originally designed for object detection in aerial images. We apply our proposed LAP image degradation pipeline on training set of DOTA-v1.0, which contains 2,806 images, to create our simulated LAP dataset. We crop each image into a size of $640\times480$. For degradation parameter, $f\in\left \{  1000, 2000, 3000, 4000\right \}$ in Hz, $A_{roll}\in\left \{ 4, 1.5, 1.0, 0.5 \right \} $ in pixel number, $A_{pitch}\in\left \{ 1, 0.5, 0.3, 0.2 \right \} $ \cite{teshima2007correction, zhu2018improved}, $\tau=3.54\times10^{-5}s$, $M=6$. We simulate four main sinusoidal components. To promise a larger degradation space, for each cropped image we multiply amplitude and frequency respectively by a vibration factor, where amplitude vibration factor satisfies Gaussian distribution with $\mu=1$, $\sigma=0.1$ and frequency vibration factor satisfies Gaussian distribution with $\mu=1$, $\sigma=0.01$. The Gaussian-Poisson noise is applied with $\sigma_{gauss}=0.01$, $\lambda_{poisson}=10^{-4}$. The maximum jitter measurement error is 20\% of current jitter offset. Overall, we obtain 20,614 train image pairs and 2,577 test image pairs in our synthetic LAP dataset.

\subsection{Implementation Details}

Our experiments are all trained and evaluated on our LAP image dataset. We set our batch size as 4 and training patch size as $128\times128$. We train our JARNet from scratch for 450k steps on a single NVIDIA GeForce RTX 4090 GPU with 24GB of memory, which takes approximately 26 hours. We apply AdamW \cite{loshchilov2018decoupled} optimizer ($\beta_{1}=0.9$, $\beta_{2}=0.999$, weight decay $10^{-3}$). Cosine learning rate strategy is applied from $3\times10^{-4}$ to $10^{-7}$. We compare our proposed method with other state-of-the-art approaches on our LAP image dataset. For a fair comparison, other methods are all trained from scratch on our LAP dataset. Except for batch size, patch size, and training steps, we follow the training protocols of each method if not specified. In testing, we use PSNR, SSIM, and gradient magnitude similarity deviation (GMSD) \cite{xue2013gradient} as our evaluation metrics with the full image size of $640\times480$.

\subsection{Comparisons with State-of-the-Art Methods}

\setlength{\tabcolsep}{2pt}
\begin{table}[h]
\small
\centering
\setlength{\tabcolsep}{0.5mm}
\begin{tabular}{c| c c c c c}
\hline
Models & $\uparrow$PSNR(dB) & $\uparrow$SSIM & $\downarrow$GMSD & $\downarrow$Params(M) \\
\hline
MIMO-UNet+ & 34.94 & 0.9127 & 0.0447 & 16.10 \\
Uformer-B & 36.08 & 0.9213 & 0.0439 & 50.88\\
HINet & 36.18 & 0.9208 & 0.0376 & 88.66\\
FNAFNet & 36.94 & 0.9311 & 0.0370 & 68.02\\
NAFNet & 36.95 & 0.9291 & 0.0366 & 67.89\\
Restormer & 37.51 & 0.9366 & 0.0344 & 26.12\\
Stripformer & \underline{37.74} & \underline{0.9389} & \underline{0.0326} & 19.71 \\
\hline
MST-L & 35.67 & 0.9155 & 0.0464 & 2.03 \\ %
MST++ & 35.65 & 0.9154 & 0.0456 & 1.33 \\ %
CST-L$^{\ast}$ & 35.16 & 0.9083 & 0.0499 & 3.00 \\ %
DAUHST-9stg & 36.08 & 0.9214 & 0.0435 & 6.15 \\ %
HDNet & 34.43 & 0.8958 & 0.0595 & 2.37 \\ %
\hline
JARNet(Ours) & \textbf{39.02} & \textbf{0.9493} & \textbf{0.0239} & 13.46 \\
\hline
\end{tabular}
\caption{Quantitative results of different RGB-based and spectral-based restoration methods and JARNet. `Params' denotes the number of parameters.}
\label{table:results}
\end{table}

We compare the proposed method on our LAP dataset with the following RGB-based methods: MIMO-UNet+ \cite{cho2021rethinking}, HINet \cite{chen2021hinet}, Uformer-B \cite{wang2022uformer}, NAFNet \cite{chen2022simple}, Restormer \cite{zamir2022restormer}, Stripformer \cite{tsai2022stripformer}, and FNAFNet \cite{mao2023intriguing}. Some spectral-based methods, such as MST \cite{mst}, MST++ \cite{mst_pp}, CST \cite{cst}, DAUHST \cite{dauhst}, and HDNet \cite{hdnet}, are also evaluated. Quantitative evaluation results in Tab. \ref{table:results} demonstrate that our proposed JARNet outperforms all other state-of-the-art methods in terms of PSNR, SSIM, and GMSD. Compared to the existing best single image restoration method, JARNet achieves 39.02dB in PSNR and 0.9493 in SSIM, which corresponds to 1.28dB improvement in PSNR, 0.0104 improvements in SSIM and 0.0087 improvements in GMSD. Fig. \ref{fig:main_result} presents visual comparison of an interior details of farmland and city buildings. Our proposed JARNet not only effectively removes jitter in LAP images but also successfully recovers more details in the contour of farmland and buildings. While other existing methods can effectively remove deformation caused by jitter, most of them fail to remove blur effect and recover sufficient details. In contrast, JARNet utilizes the prior knowledge of jitter to make pre-correction in the first stage and reduces the learning difficulty for subsequent restoration network so as to pay more attention to image details.

\section{Ablation Studies}

\subsection{Effectiveness of Pre-Correction}

\setlength{\tabcolsep}{2pt}
\begin{table}[h]
\small
\centering
\setlength{\tabcolsep}{0.5mm}
\begin{tabular}{c| c c c c}
\hline
Models & $\uparrow$PSNR(dB) & $\uparrow$SSIM & $\downarrow$GMSD &  $\downarrow$GMACs\\
\hline
MIMO-UNet+$^{\ast}$ & 36.07 & 0.9402 & 0.0293 & 154.3 \\
Uformer-B$^{\ast}$ & 37.93 & 0.9431 & 0.0286 & 85.75 \\
HINet$^{\ast}$ & 37.97 & 0.9425 & 0.0275 & 170.3 \\
FNAFNet$^{\ast}$ & 37.70 & 0.9404 & 0.0296 & 63.34 \\
NAFNet$^{\ast}$ & 38.01 & 0.9428 & 0.0279 & 63.18 \\
Restormer$^{\ast}$ & 38.46 & 0.9472 & 0.0263 & 140.8 \\
Stripformer$^{\ast}$ & \underline{38.72} & \textbf{0.9500} & \underline{0.0247} & 170.4 \\
JARNet(Ours) & \textbf{39.02} & \underline{0.9493} & \textbf{0.0239} & 85.71 \\
\hline
\end{tabular}
\caption{Quantitative results of different enhanced RGB-based restoration methods. All methods denoted `$*$' are enhanced by pre-correction in the first stage. We evaluate GMACs with input tensor shape of $(1, 1, 256, 256)$.}
\label{table:warp_results}
\end{table}

To validate the effectiveness of pre-correction, we utilize jitter prior to enhance other methods in a two-stage restoration manner as shown in Fig. \ref{fig:pipeline} (c). Tab. \ref{table:warp_results} shows that performances of all methods enhanced by jitter prior are improved, compared with Tab. \ref{table:results}. Notably, Stripformer$^{*}$ achieves a slightly higher SSIM compared with our method, while JARNet maintains the best PSNR, GMSD metric, and relatively acceptable amount of GMACs among all methods. The pre-correction in the first stage effectively reduces learning difficulty of restoration network in the second stage, leading to enhanced performance for mainstream methods.

\begin{table}[h]
\centering
\begin{tabular}{c|ccc|ccc}
\hline
No. & Freq & CoA & OFC & $\uparrow$PSNR & $\uparrow$SSIM & $\downarrow$Params(M) \\
\hline
1 & & & & 38.17 & 0.9444 & 13.80 \\
2 & $\checkmark$ & & & 37.98 & 0.9428 & 13.94 \\
3 & & $\checkmark$ & & 38.50 & 0.9478 & 12.26 \\
4 & & & $\checkmark$ & 38.56 & 0.9463 & 14.87 \\
5 & $\checkmark$ & $\checkmark$ & & \underline{38.69} & \underline{0.9492} & 12.39 \\
6 & $\checkmark$ & & $\checkmark$ & 38.44 & 0.9456 & 15.01 \\
7 & & $\checkmark$ & $\checkmark$ & \underline{38.69} & 0.9471 & 13.33 \\
8(Ours) & $\checkmark$ & $\checkmark$ & $\checkmark$ & \textbf{39.02} & \textbf{0.9493} & 13.46 \\
\hline
\end{tabular}
\caption{Ablation of frequency branch, CoA block, and OFC block in JARNet. `Freq' denotes frequency branch.}
\label{tab:component_comparison}
\end{table}

\subsection{Effectiveness of Components in JARNet}

In this section, we verify the effectiveness of frequency branch, CoA block, and OFC block in JARNet by conducting 7 extra experiments.
As shown in Tab. \ref{tab:component_comparison}, when we add CoA block alone in No.3, we observe a significant performance gain of 0.33dB in PSNR compared with No.1 baseline. This improvement indicates that the CoA block is effective in extracting orthogonal jitter state. Additionally, the number of parameters reduces by  11.2\% approximately, indicating the efficiency of incorporating the CoA block in our JARNet.
When we add frequency branch in No.2, the performance has a decline of 0.19dB in PSNR, compared with No.1 baseline. Nevertheless, by incorporating the global information extracted by frequency branch and jitter state extracted by CoA block, No.5 achieves better performance compared with No.3. So we apply frequency branch and CoA block simultaneously.
Similarly, when we add OFC block in No.4, we achieve a performance gain of 0.39dB in PSNR, compared with No.1 baseline. This improvement indicates that refined optical flow effectively pre-corrects degraded LAP image. For the best restoration performance, we apply all three components in No.8 JARNet.

\begin{figure}
\centering
\includegraphics[width=0.47\textwidth]{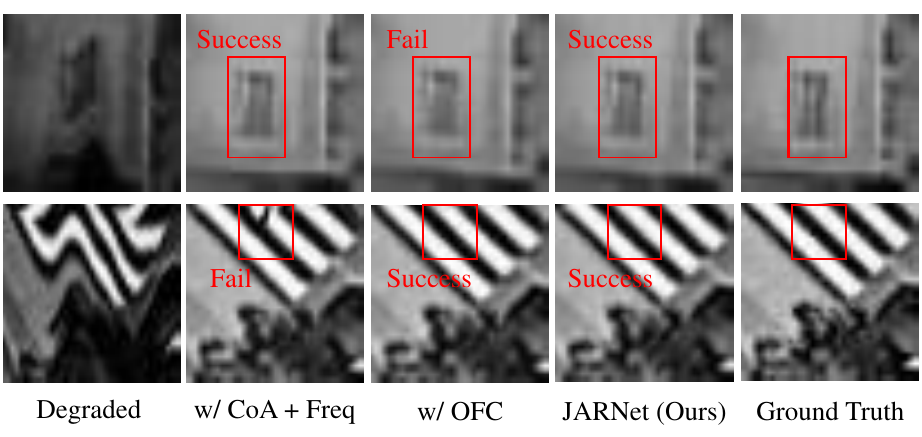}
\caption{Visual comparison of different components.}
\label{fig:component_ablation}
\end{figure}

We further analyze visual results of different combinations of components. As shown in Fig. \ref{fig:component_ablation}, when the frequency branch and CoA block are combined, the vertical details in the window are partly restored, which indicates that the CoA block, along with the frequency domain branch, has capabilities to capture jitter state and high-frequency features. OFC block helps reconstruct zebra crossing correctly on the edge of the degraded
\begin{table}[h]
\centering
\begin{tabular}{lcccc}
\hline
& NAFNet$^{\ast}$ & Restormer$^{\ast}$ & Stripformer$^{\ast}$ & JARNet(Ours) \\
\hline
PSNR & 36.64 & 37.06 & 37.52 & 38.50 \\
SSIM & 0.9270 & 0.9338 & 0.9403 & 0.9465 \\
\hline
\end{tabular}
\caption{Results on LAP dataset without CDSM.}
\label{tab:CDSM_ablation}
\end{table}
image. In situations where pre-corrected images have black edges, architectures without OFC block fail to handle the edge details correctly. However, the application of the OFC block helps alleviate the influence of these black edges, resulting in better restoration of edge details. Our JARNet combines advantages of all three components and achieves high-quality restoration results.

\subsection{Effectiveness of CDSM}

In order to verify CDSM enhances the restoration effect of the LAP dataset. We remove CDSM and generate a new CDSM-free LAP dataset on which we compare different enhanced state-of-the-art methods. As shown in Tab. \ref{tab:CDSM_ablation}, other enhanced methods suffer from a sharp performance decline compared with results in Tab. \ref{table:warp_results}, even worse than non-enhanced versions in Tab. \ref{table:results}. Without CDSM, the noisy jitter curve is not smoothed. Therefore, pre-correction in the first stage does not work. However, JARNet only shows a slight decrease in performance, because the OFC block in the first stage compensates part of smoothing effect from CDSM. This fully demonstrates the importance of CDSM in LAP image restoration pipeline. CDSM helps smooth the noisy jitter curve and obtain better pre-correction results.

\section{Conclusion}

In order to restore degraded LAP image caused by camera jitter and overcome the obstacle of data availability, we proposed a CDSM-based LAP image degradation pipeline and created a LAP dataset. Then we presented the first jitter-aware restoration network in a two-stage restoration manner. In the first stage, we utilized optical flow refined by OFC block to warp the degraded LAP image. In the second stage, we incorporated CoA block and frequency branch in SFRes block to realize jitter-aware character. CoA block captures jitter state in orthogonal direction, while frequency branch extracts both low and high-frequency jitter. Extensive experiments demonstrate that our approach performs favorably against state-of-the-art methods qualitatively and quantitatively on our LAP dataset.

\section{Acknowledgments}
This project is supported by National Natural Science Foundation of China (No. 62275229). We thank Meijuan Bian and Weige Lyu from the facility platform of optical engineering of Zhejiang University for instrument support. 

\section{Appendix}

\subsection{Approximation of Optical Flow}
In this section, we discuss the reason why we make an approximation between optical flow and the inverse of jitter map. For simplicity we only consider distortion. The process of deforming a clean image is shown in Eq. \ref{approximation1}:
\begin{equation}
\begin{split}
I^{lq}(h+\Delta h, w+\Delta w) =\boldsymbol G(I^{gt}(h,w), J(h,w)),
\label{approximation1}
\end{split}
\end{equation}
where $h$ and $w$ represent pixel position in the image, and $J$ is a jitter map. $\Delta h$ and $\Delta w$ is pixel displacement at original position $(h, w)$. We pre-correct the deformed image precisely in Eq. \ref{approximation2}:
\begin{equation}
\begin{split}
I^{warp}(h^{'}, w^{'}) &= \boldsymbol G(I^{lq}(h + \Delta h, w + \Delta w), -J(h,w)),
\label{approximation2}
\end{split}
\end{equation}
where $h^{'}$ and $w^{'}$ represent pixel position in the pre-corrected image. Theoretically, we need jitter data at the original position $(h, w)$. However, we only have deformed image $I^{lq}(h + \Delta h, w + \Delta w)$ and measured jitter data $J(h+\Delta h, w+\Delta w)$. We do not know where the original position $(h, w)$ is. On the other hand, $\Delta h$ and $\Delta w$ are probably not integers. In practice, $I^{lq}$ is processed by bilinear interpolation algorithm during the process of grid sample. 

To make Eq. \ref{approximation2} computable, we assume the jitter data at new position $(h+\Delta h, w+\Delta w)$ is approximately equaled to that in original position $(h, w)$, as shown in Eq. \ref{approximation3}:
\begin{equation}
\begin{split}
J(h,w) = J(h + \Delta h, w) \approx J(h + \Delta h, w + \Delta w),
\label{approximation3}
\end{split}
\end{equation}
For height direction, the jitter state keeps the same, because the pixels along the height direction are imaged at the same time. We make an approximation in the width direction, because jitter offsets in the along-track direction are much smaller than that in the cross-track direction \cite{wang2017image}. So we treat the inverse of jitter map $-J(h + \Delta h, w + \Delta w)$ as optical flow for pre-correction.

\begin{figure}[h]
\centering
\includegraphics[width=0.47\textwidth]{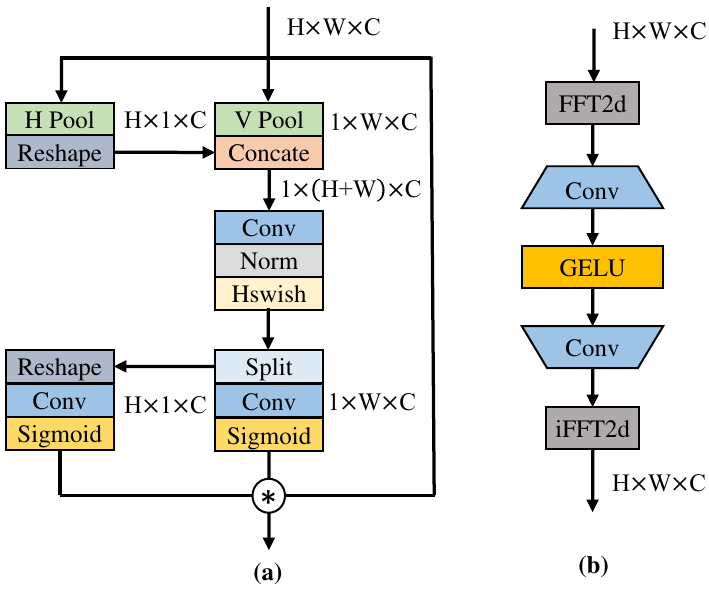}
\caption{(a) Details of CoA block \cite{hou2021coordinate}. `H Pool' means global pooling operation along the horizontal direction. `V Pool' means global pooling operation along the vertical direction. `Hswish' means hswish activation function \cite{RamachandranZL18}. (b) Details of frequency branch \cite{mao2023intriguing}.
}
\label{fig:coafft}
\end{figure}

\begin{figure}[h]
\centering
\includegraphics[width=0.47\textwidth]{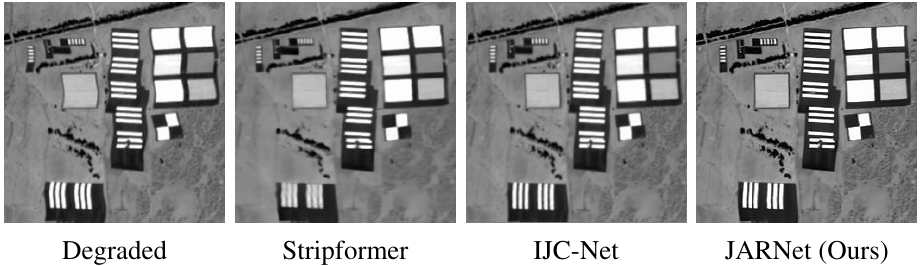}
\caption{Visual comparison of real LAP image `Lijiang' from Yaogan-26 satellite.
Please zoom in for a better view.}
\label{fig:realLAP}
\end{figure}

\begin{figure*}[h!]
    \centering
    \includegraphics[width=\textwidth]{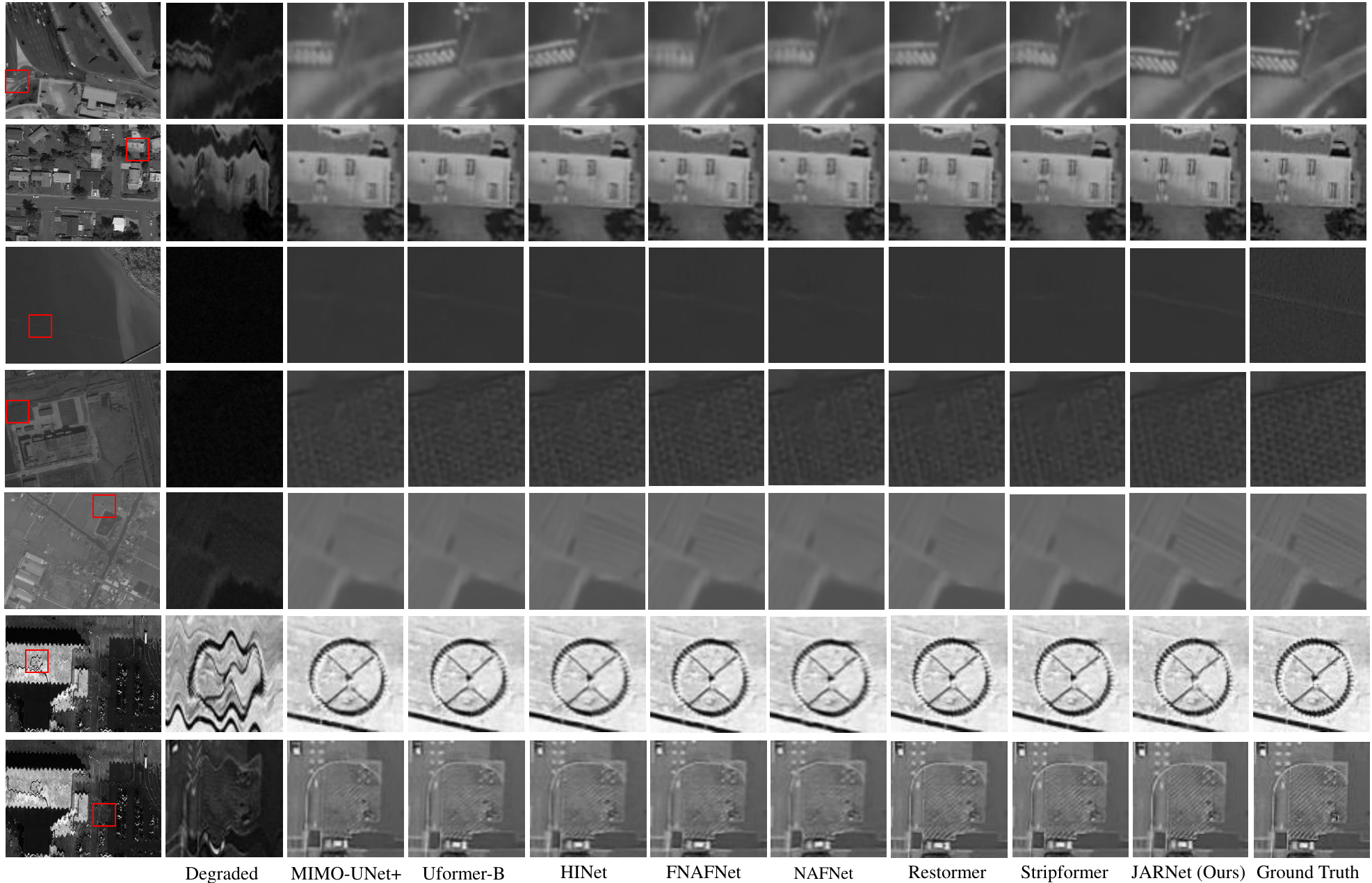}
    \caption{More visual comparisons of different restoration methods on our LAP dataset.}
    \label{fig:sub_visual}
\end{figure*}

\begin{figure}[h!]
    \centering
    \includegraphics[width=0.47\textwidth]{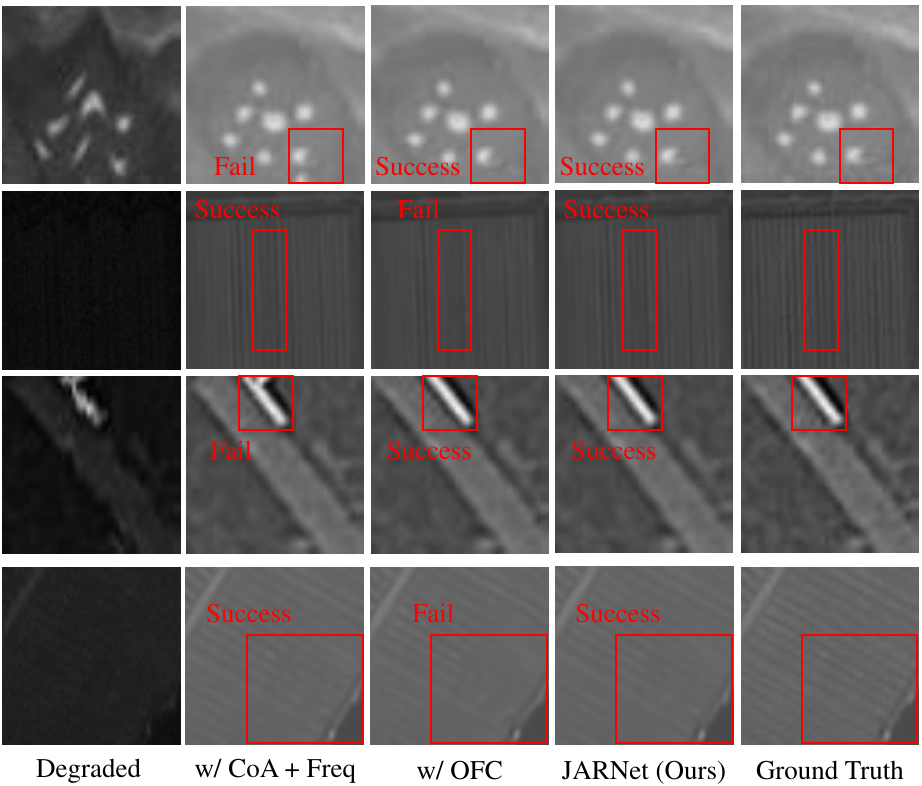}
    \caption{More visual comparisons of different components.}
    \label{fig:sub_ablation}
\end{figure}

\subsection{More Architecture Details of JARNet}

We adopt NAFNet \cite{chen2022simple} as the backbone of JARNet, with 4 levels of encoder and decoder. We set 32 as the width of JARNet, with 36 SFRes blocks in total. Each level of encoder and decoder contains 4 SFRes blocks, and the middle bottleneck also contains 4 SFRes blocks.

We demonstrate the architecture details of the CoA block and frequency branch in Fig. \ref{fig:coafft}. The CoA block manipulates attention maps in both horizontal and vertical directions. Feature map with the shape of $(H, W, C)$ is applied element-wise multiplication respectively with two attention maps by the broadcast mechanism in Python.

It is worth noting that different levels of encoder and decoder extract information at varying levels. To control the receptive field, we employ a window partition strategy \cite{mao2023intriguing} in the frequency branch. This strategy is taken because shallow levels focus on extracting local degradation. On the other hand, deep levels are responsible for detecting global jitter states. In practice, we set the window size in the first level as 64, which means only a $64\times64$ spatial region is considered when extracting frequency domain information. Window partition in other levels is disabled, allowing for feature extraction in the global context.

\subsection{Real-World LAP Image Restoration Results}

We conducted experiments on real-world LAP images from
Yaogan-26 satellite \cite{wang2017image} and show results in Fig. \ref{fig:realLAP}. We did the same pre-processing for degraded image before implementing Stripformer and JARNet. Compared to Stripformer and IJC-Net \cite{zhaoxiang2019attitude}, JARNet achieves the best visual result, where most of the distortion and blur are removed.

\subsection{More Visual Results of Restored LAP Images}

As shown in Fig. \ref{fig:sub_visual}, we demonstrate more visual comparisons of different restoration methods through 7 groups. Fig. \ref{fig:sub_ablation} shows more visual comparisons of different components in our JARNet through 4 groups.

\subsection{Recovery Performance on More Remote Datasets}

In order to validate the generalization ability of JARNet, we utilize the model trained on our original LAP dataset from DOTA-v1.0 to restore other remote datasets simulated by our degradation pipeline without finetune.

We choose AID and PatternNet as our samples. AID \cite{xia2017aid} is a large-scale aerial image dataset by collecting sample images from Google Earth imagery. AID dataset contains 10,000 images of size $600 \times 600$ within 30 classes. PatternNet \cite{zhou2018patternnet} is a large-scale high-resolution dataset for remote sensing image retrieval. There are 38 classes and each class has 800 images of size $256 \times 256$.

We make two new LAP datasets (abbreviated as the original name of the dataset) based on AID and PatternNet through our LAP degradation pipeline, respectively. We crop images in AID into a size of $480 \times 360$, while we keep the original resolution in PatternNet. All simulated image pairs are utilized for test. We acquire 10,000 image pairs in AID and 30.400 image pairs in PatternNet.

\begin{table}[h!]
\centering
\begin{tabular}{c|cc|cc}
\hline
Models & $\uparrow$PSNR(dB) & $\uparrow$SSIM & $\uparrow$PSNR(dB) & $\uparrow$SSIM\\
\hline
NAFNet & 28.66 & 0.8798 & 30.46 & 0.8727 \\
Restormer & 32.79 & 0.9038 & 31.22 & 0.8863 \\
Stripformer & 32.92 & 0.9052 & 31.28 & 0.8869 \\
\hline
NAFNet$^{\ast}$ & 32.06 & 0.9048 & 31.55 & 0.8981 \\
Restormer$^{\ast}$ & 33.47 & 0.9204 & 32.15 & 0.9072 \\
Stripformer$^{\ast}$ & \underline{33.81} & \textbf{0.9247} & \underline{32.39} & \underline{0.9107} \\
\hline
JARNet & \textbf{34.12} & \underline{0.9241} & \textbf{33.40} & \textbf{0.9159} \\
\hline
\end{tabular}
\caption{Quantitative results on AID \cite{xia2017aid} dataset ($2^{nd}$ and $3^{rd}$ column) and PatternNet \cite{zhou2018patternnet} dataset ($4^{th}$ and $5^{th}$ column).}
\label{tab:AID_PatternNet}
\end{table}

As shown in Tab. \ref{tab:AID_PatternNet}, our JARNet keeps a good performance while transferring to other datasets without finetune. Compared with other state-of-the-art methods, our JARNet keeps the performance advantage on AID and PatternNet, showing the great generalization ability of JARNet.

\bibliography{aaai24}

\end{document}